\documentclass[10pt,twocolumn,letterpaper]{article}

\usepackage{iccv}      

\definecolor{iccvblue}{rgb}{0.21,0.49,0.74}
\usepackage[pagebackref,breaklinks,colorlinks,allcolors=iccvblue]{hyperref}

\def\paperID{*****} 
\def\confName{ICCV}
\def\confYear{2025}

\title{Improving VQA Reliability: A Dual-Assessment Approach with Self-Reflection and Cross-Model Verification}

\author{
Xixian Wu\\
Bilibili Inc.\\
Shanghai, China\\
{\tt\small wuxixian@bilibili.com}
\and
Yang Ou\\
Bilibili Inc.\\
Shanghai, China\\
{\tt\small ouyang10@bilibili.com}
\and
Pengchao Tian\\
Bilibili Inc.\\
Shanghai, China\\
{\tt\small tianpengchao@bilibili.com}
\and
Zian Yang\\
Bilibili Inc.\\
Shanghai, China\\
{\tt\small yangzian@bilibili.com}
\and
Jielei Zhang\\
Bilibili Inc.\\
Shanghai, China\\
{\tt\small zhangjielei@bilibili.com}
\and
Peiyi Li\\
Bilibili Inc.\\
Shanghai, China\\
{\tt\small lipeiyi@bilibili.com}
\and
Longwen Gao\\
Bilibili Inc.\\
Shanghai, China\\
{\tt\small gaolongwen@bilibili.com}
}

\begin{document}
\maketitle
\begin{abstract}
Vision-language models (VLMs) have demonstrated significant potential in Visual Question Answering (VQA). However, the susceptibility of VLMs to hallucinations can lead to overconfident yet incorrect answers, severely undermining answer reliability. To address this, we propose Dual-Assessment for VLM Reliability (DAVR), a novel framework that integrates Self-Reflection and Cross-Model Verification for comprehensive uncertainty estimation. The DAVR framework features a dual-pathway architecture: one pathway leverages dual selector modules to assess response reliability by fusing VLM latent features with QA embeddings, while the other deploys external reference models for factual cross-checking to mitigate hallucinations. Evaluated in the Reliable VQA Challenge at ICCV-CLVL 2025, DAVR achieves a leading $\Phi_{100}$ score of 39.64 and a 100-AUC of 97.22, securing first place and demonstrating its effectiveness in enhancing the trustworthiness of VLM responses.
\end{abstract}    
\section{Introduction}
\label{sec:intro}
Visual Question Answering (VQA) is a core multimodal task that evaluates a model’s capacity for integrating visual comprehension with linguistic reasoning. While Vision-Language Models (VLMs) have achieved impressive benchmark scores, their practical deployment—especially in safety-sensitive applications—is hampered by a critical lack of answer reliability. This issue stems largely from the models' opaque reasoning and propensity for overconfident hallucinations, highlighting an urgent need for effective uncertainty quantification in VLM responses.

One way to address this issue is to formulate the problem as selective prediction. In this paradigm, a model is not only required to produce an output but also to assesse the reliability of its prediction and abstains when uncertain. In the context of the Reliable VQA Challenge, a sub-challenge of the ICCV-CLVL 2025 workshop, participants are required to not only generate answers to VQA tasks but also provide a confidence score for each response. The evaluation additionally requires specifying a global abstention threshold, under which responses are regarded as uncertain and consequently treated as abstentions.

Recently, several studies have been proposed to solve such problem, \cite{whitehead2022reliable} explores the effectiveness of several selection functions including MaxProb, Calibration and Multimodal selection function. \cite{dancette2023improving} propose Learning from Your Peers (LYP) approach for training multimodal selection functions for making abstention decisions. Other efforts~\cite{srinivasan2024selective, chen2023adaptation} have attempted to enable VLMs to assess the reliability of their own answers. However, due to the intrinsic hallucination problem of VLMs, the model may still exhibit overconfidence in its erroneous predictions.

In this work, we introduce the Dual-Assessment for VLM Reliability (DAVR) framework. DAVR combines model self-reflection with cross-model verification to first assess answer reliability and then leverage inter-model consensus to reduce overconfidence and hallucinations. In summary, our contribution can be expressed as follows:
\begin{itemize}
    \item We introduce a cross-model verification mechanism on top of model self-reflection, effectively mitigating the intrinsic hallucination problem of VLMs.
    \item We leverage an ensemble of multiple models to enhance the estimation of prediction reliability, resulting in substantial improvements.
    \item Extensive evaluation demonstrates that our approach achieves state-of-the-art performance on the Reliable VQA Challenge leaderboard.
\end{itemize}

\begin{figure*}[htbp]
    \centering
    \includegraphics[width=\textwidth]{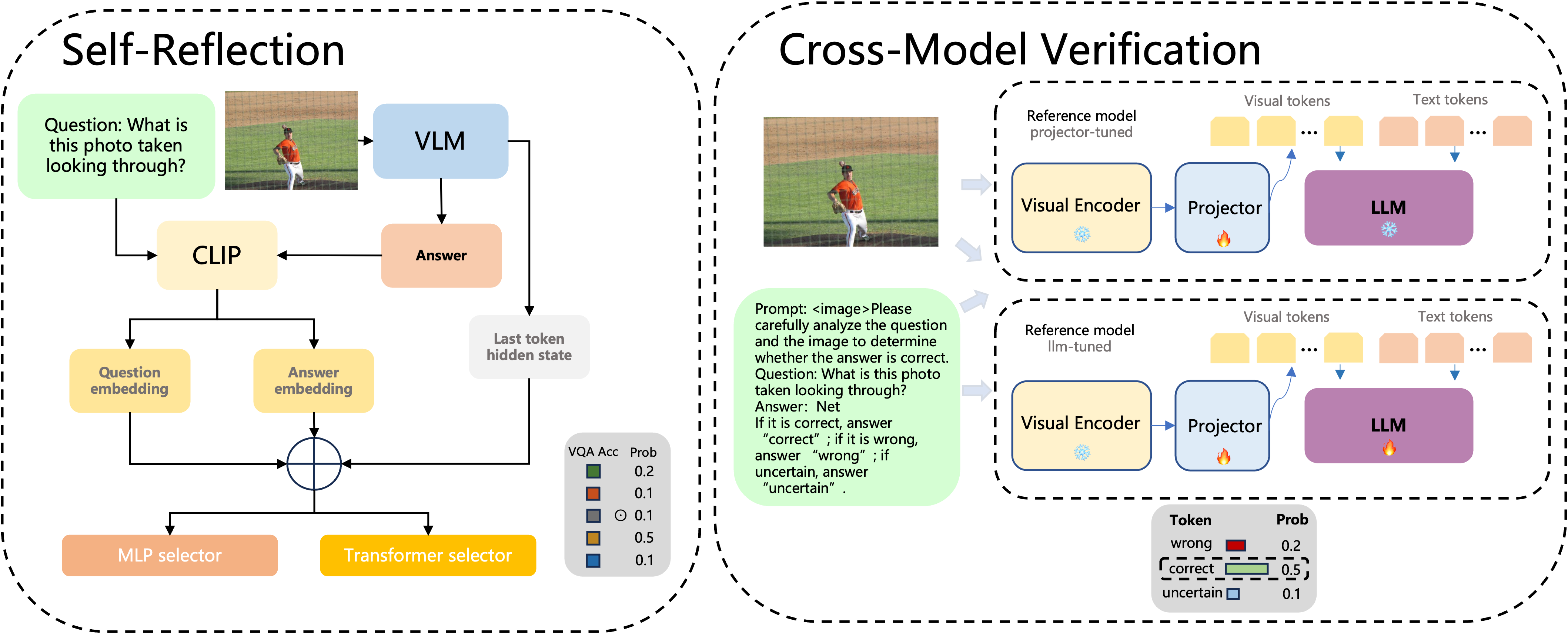}
    \caption{Overall framework of the DAVR.}
    \label{fig:DAVR}
\end{figure*}

\section{Method}
The overall framework of our approach is illustrated in Figure~\ref{fig:DAVR}.  
We first assess the reliability of the VLM’s outputs using its internal representations.  
Then, we validate the answers with external reference models and generate robust predictions through model ensembling.

\subsection{Self-Reflection}
To assess the model’s uncertainty in its own predictions, we extract the last-layer hidden states from the VLM during inference for each VQA image-text pair. These representations inherently encode the rich multimodal interactions between visual and linguistic inputs~\cite{zhang2024visuallygroundedlanguagemodelsbad}. In addition, we extract textual embeddings of both the question and the model-generated answer using CLIP, providing complementary semantic cues to the multimodal representations.
Considering that the VQA accuracy is a discrete value, we formulate the problem as a classification task, which tends to be more stable and easier to optimize than a regression formulation. 

Formally, we consider a dataset $\mathcal{D} = \{(x_i, y_i)\}_{i=1}^{N}$, where $y_i$ represents the discretized VQA accuracy serving as a categorical label. Each input $x_i$ consists of the VLM's hidden state, the embedding of the question, and that of the answer, i.e.,$x_i = [h_i; q_i; a_i],$ where $h_i$ denotes the hidden state representation, $q_i$ the textual feature of the question, and $a_i$ the textual feature of the answer.

The learning objective is formulated as a standard multi-class classification problem. Given the categorical label $y_i$, the model is trained to learn the mapping from $x_i$ to $y_i$ by minimizing the cross-entropy loss. Formally, the loss function is defined as: 
\begin{equation}
    \mathcal{L}_{\text{CE}}(\theta) = - \frac{1}{N} \sum_{i=1}^{N} \sum_{c=0}^{C-1} \mathbb{I}(y_i = c) \log p_{\theta}(y_i = c \mid x_i),
\end{equation}
with $C$ denoting the total number of classes and $\mathbb{I}(\cdot)$ representing the indicator function. This objective encourages the predicted probability distribution $p_{\theta}(y_i \mid x_i)$ to be closely aligned with the ground-truth label distribution.

In practice, due to the relatively low proportion of negative samples, we adopt the focal loss to address this imbalance issue and improve the model’s learning on hard examples. The focal loss is formulated as follows:
\begin{align}
\mathcal{L}_{\text{FL}}(\theta) 
&= - \frac{1}{N} \sum_{i=1}^{N} \sum_{c=0}^{C-1} \alpha_c \, (1 - p_{\theta}(y_i = c \mid x_i))^\gamma \notag\\
&\quad \times \mathbb{I}(y_i = c) \, \log p_{\theta}(y_i = c \mid x_i),
\end{align}
where $\alpha_c \in [0,1]$ is a weighting factor for class $c$ to address class imbalance, and $\gamma > 0$ is the focusing parameter that emphasizes hard-to-classify examples.  

At inference time, we estimate the confidence score $\hat{s}_i$ by taking the expectation of the discretized accuracy values with respect to the predicted categorical distribution:
\begin{equation}
    \hat{s}_i = \sum_{c=0}^{C-1} a_c \, p_{\theta}(y_i = c \mid x_i),
\end{equation}
where $a_c$ denotes the VQA accuracy associated with class $c$. This formulation allows the model to output a continuous confidence estimate based on the learned discrete probability distribution.

Based on the above, we independently train two models, one using an MLP architecture and the other using a Transformer architecture. For each sample, the resulting confidence scores are denoted as $\hat{s}_i^{\text{MLP}}$ and $\hat{s}_i^{\text{T}}$, respectively.

\begin{table*}[ht]
\centering
\caption{Comparison with other methods.}
\label{tab:main_results}  
\begin{tabular}{lccccccc}
\hline
\textbf{Team} & \textbf{Accuracy} & \textbf{\(\Phi_{100}\)} & \textbf{100-AUC} & \textbf{Cov@0.5\%} & \textbf{Cov@1\% (\%)} & \textbf{Cov@2\%} & \textbf{Cov@5\%} \\

\hline
jokur (Test-standard)        & \underline{84.05} & \underline{32.06} & \underline{96.85} & \underline{40.67} & \underline{49.30} & \underline{59.15} & \underline{75.30} \\
TIMM (Test-standard)     & 83.91 & 31.31 & 96.63 & 37.93 & 46.46 & 56.38 & 72.86 \\
Freedom (Test-standard)     & 82.99 & 22.73 & 96.72 & 29.21 & 44.71 & 58.84 & 76.65 \\
Baseline (Test-standard)     & 81.35 & 23.00 & 96.09 & 26.98 & 39.75 & 52.88 & 71.78 \\
\hline
Ours (Test-dev)      & 83.99 & 37.52 & 97.15 & 44.56 & 53.03 & 63.49 & 77.97 \\
Ours (Test-standard) & \textbf{84.11} & \textbf{39.64} & \textbf{97.22} & \textbf{45.67} & \textbf{54.12} & \textbf{63.71} & \textbf{78.55} \\
\hline
\end{tabular}
\end{table*}

\begin{table}[ht]
\centering
\caption{performance on the VQAV2 validation set.}
\label{tab:vlm_performance}  
\begin{tabular}{l|c}
\hline
\textbf{Model} & \textbf{Accuracy (\%)} \\
\hline
QwenVL2-7B-Inst             & \underline{82.97} \\
QwenVL2-7B-Inst CoT         & 65.91 \\
InternVL3\_5-8B             & 74.08 \\
InternVL3\_5-8B-Inst        & 78.02 \\
InternVL3\_5-GPT-OSS-20B    & 80.06 \\
InternVL3\_8B-Inst          & 79.79 \\
QwenVL2.5-7B-Inst           & 76.23 \\
\hline
QwenVL2-7B-Inst (TTA)       & \textbf{83.14} \\
\hline
\end{tabular}
\label{tab:model_comparison}
\end{table}



\begin{table}[ht]
\centering
\caption{Ensemble of Models.}
\label{tab:model_ensemble}  
\begin{tabular}{l|ccc}
\hline
\textbf{Team} & \textbf{\(\Phi_{100}\)} & \textbf{100-AUC} & \textbf{Cov@0.5\%} \\

\hline
MLP   & {30.91} & {96.67} & {37.79} \\
MLP+T & {31.52} & {96.7} & {37.94}\\
MLP+T+P & {36.31} & {97.03} & {43.13}\\
MLP+T+P+L & \textbf{39.64} & \textbf{97.22} & \textbf{45.67}\\
\hline
\end{tabular}
\end{table}

\subsection{Cross-Model Verification}
The selector heavily relies on the quality of the VLM hidden states. To mitigate this limitation, we introduce external reference models to verify the answers generated by the VLM. 

Specifically, given the original VQA pair and the answer generated by the VLM, we first construct a verification dataset by categorizing VLM-generated answers based on their VQA accuracy: answers with an accuracy of 0 are labeled as wrong, those with accuracy greater than 0 but less than or equal to 0.66 were labeled as uncertain, and those greater than 0.66 as correct. We use Qwen2.5-VL-7B-Inst to assess whether the answer is \emph{correct}, \emph{wrong}, or \emph{uncertain}. We derive a confidence score from the probability that Qwen2.5-VL generates the word \texttt{"correct"}:
\begin{equation}
    \hat{s}_i^{\text{VLM}} = p_{\text{Qwen}}(\text{"correct"} \mid x_i, \text{VLM answer}),
\end{equation}
where $p_{\text{Qwen}}(\cdot)$ is the predicted probability assigned by the external reference model to its first output token.

To enhance model reliability,  We perform supervised fine-tuning (SFT) of the external reference model Qwen2.5-VL on the labeled dataset. Through differentiated fine-tuning strategies, we derive two distinctively adapted versions: one involves fine-tuning only the multi-modal projector (yielding the only-Projector-tuned model), while the other fine-tunes both the projector and the language model (yielding the language-tuned model). All fine-tuning is conducted for a single epoch. Subsequently, inference is performed with these models to obtain confidence scores for each sample, denoted as $\hat{s}_i^{\text{VLM-P}}$ and $\hat{s}_i^{\text{VLM-L}}$, respectively. These scores represent the estimated probability that a given VLM-generated answer is correct.

To this end, we progressively aggregate the outputs from all models to obtain the final confidence score, which is expressed as follows:
\begin{align}
    \hat{s}_i^{\text{MLP+T}} &= 0.5 \hat{s}_i^{\text{MLP}} + 0.5 \hat{s}_i^{\text{T}}, \\
    \hat{s}_i^{\text{MLP+T+P}} &= 0.5 \hat{s}_i^{\text{MLP+T}} + 0.5 \hat{s}_i^{\text{VLM-P}}, \\
    \hat{s}_i^{\text{MLP+T+P+L}} &= 0.5 \hat{s}_i^{\text{MLP+T+P}} + 0.5 \hat{s}_i^{\text{VLM-L}}.
\end{align}

\section{Experiments}
\subsection{Implement Details}
We employ QwenVL2-7B-Inst for inference due to its strong performance on the VQAv2 dataset. To further enhance performance, we apply test-time augmentation (TTA) by increasing the temperature and aggregating multiple inference results, which leads to a slight improvement in overall accuracy.

For the MLP selector, the question and answer features are each processed through one fully connected layer, while the hidden state features pass through two fully connected layers. The resulting features are concatenated and fed into three fully connected layers for classification. In contrast, for the Transformer selector, a two-layer Transformer encoder is used to model interactions among the three features, and classification is performed based on the hidden state token. The checkpoint achieving the best performance on the validation set is used for testing.

The abstention threshold is determined by sweeping across different threshold values and selecting the one that yields the best performance on the validation set.
\subsection{Main Results}
The final results are presented in the table~\ref{tab:main_results}, including only the public leaderboard entries and the performance of our method. While the accuracy is comparable to existing approaches, our method consistently outperforms them across all other metrics, achieving a \(\Phi_{100}\) of 39.64, a 100-AUC of 97.22, and a Cov@0.5\% of 45.67, among others. Especially, in terms of \(\Phi_{100}\), our method outperforms the second-best approach by 7.58.
\subsection{Ablation Study}

\textbf{Performance of VLMs on the VQAV2 dataset: }From Table~\ref{tab:vlm_performance}, we observe that Qwen2 achieves the best performance on the VQAv2 dataset, even surpassing its subsequent versions. Moreover, applying TTA can slightly enhance the performance of the VLM.

\noindent\textbf{Ensemble of Models: } We report the performance of \({\text{MLP+T}}\), \({\text{MLP+T+P}}\), and \({\text{MLP+T+P+L}}\) in Table~\ref{tab:model_ensemble}. The results indicate that each model contributes positively to the prediction, demonstrating the effectiveness of our approach.

\section{Conclusion}
In this work, we present a novel selective prediction framework for vision-language models that combines model self-reflection with cross-model verification. By enabling models to internally assess the reliability of their own predictions and to leverage consensus across multiple models, our approach effectively mitigates overconfident and hallucinated responses, a major challenge in deploying VLMs in safety-critical applications. Extensive experiments on the Reliable VQA Challenge demonstrate that our method not only improves the trustworthiness of predictions but also establishes a new state-of-the-art performance on the leaderboard. In the future, we plan to extend this framework to broader multi-modal reasoning tasks and explore more efficient mechanisms for model collaboration and uncertainty estimation.

{
    \small
    \bibliographystyle{ieeenat_fullname}
    \bibliography{main}
}

\end{document}